\documentclass[conference]{IEEEtran}
\IEEEoverridecommandlockouts

\usepackage{cite}
\usepackage{amsmath,amssymb,amsfonts}
\usepackage{algorithmic}
\usepackage{graphicx}
\usepackage{textcomp}
\usepackage{xcolor}
\def\BibTeX{{\rm B\kern-.05em{\sc i\kern-.025em b}\kern-.08em
    T\kern-.1667em\lower.7ex\hbox{E}\kern-.125emX}}

\usepackage{hyperref}

\usepackage{natbib}
\usepackage{multirow}
\usepackage{comment}
\usepackage{amsmath}  
\usepackage{amsfonts}  
\usepackage{amssymb}  
\usepackage{booktabs}  
\usepackage{xcolor}  

\begin{document}


\title{ Diffusion Is Your Friend in Show, Suggest and Tell }



\author{\IEEEauthorblockN{Jia Cheng Hu$^1$, Roberto Cavicchioli$^2$, Alessandro Capotondi$^3$}

\IEEEauthorblockA{
\textit{$^{1,3}$Department of Physics, Informatics and Mathematics}
\textit{$^2$Department of Communication and Economics}\\
\textit{University of Modena and Reggio Emilia}\\
\{name.surname\}@unimore.it}
$^1$https://orcid.org/0009-0008-1611-966X $^2$https://orcid.org/0000-0003-0166-0898 $^3$https://orcid.org/0000-0001-8705-0761
}


\maketitle

\begin{abstract}
    Diffusion Denoising models demonstrated impressive results across generative Computer Vision tasks, but they still fail to outperform standard autoregressive solutions in the discrete domain, and only match them at best. In this work, we propose a different paradigm by adopting diffusion models to provide suggestions to the autoregressive generation rather than replacing them. By doing so, we combine the bidirectional and refining capabilities of the former with the strong linguistic structure provided by the latter. To showcase its effectiveness, we present Show, Suggest and Tell (SST), which achieves State-of-the-Art results on COCO, among models in a similar setting. In particular, SST achieves 125.1 CIDEr-D on the COCO dataset without Reinforcement Learning, outperforming both autoregressive and diffusion model State-of-the-Art results by 1.5 and 2.5 points. On top of the strong results, we performed extensive experiments to validate the proposal and analyze the impact of the suggestion module. 
    Results demonstrate a positive correlation between suggestion and caption quality, overall indicating a currently underexplored but promising research direction. Code will be available at: \url{https://github.com/jchenghu/show\_suggest\_tell}.
\end{abstract}


\section{Introduction}
\label{sec:intro}

Denoising Diffusion Models \cite{sohl2015deep, ho2020denoising} exhibited impressive results across many Computer Vision applications, such as Image Generation, Image Processing, and 3D Reconstruction. Unfortunately, they fail to deliver similar improvements in areas involving text, such as Image Captioning. Diffusion models in sequence modeling can be seen as the latest trend in an ongoing debate between non-autoregressive (NAR) and autoregressive (AR) models, whose origin can be traced back nearly a decade ago.  While the latest discrete diffusion proposals \cite{chen2022analog, wang2024decap, yang2024captioner} achieve promising results, they still fall short in terms of accuracy compared to the autoregressive counterparts. The latter models are still the dominant and preferred approach. Diffusion models are usually preferred for the inference time reduction \cite{wu2023ar}, yet, such an advantage is jeopardized by the branch of research that focuses on designing faster autoregressive models \cite{sun2023retentive, hu2024shifted}. This highlights the need to find alternative purposes for diffusion models. For instance, one that supports the autoregressive generation rather than challenging the status quo. In this 
\begin{figure}[ht]
  \centering
\includegraphics[width=0.4\textwidth]{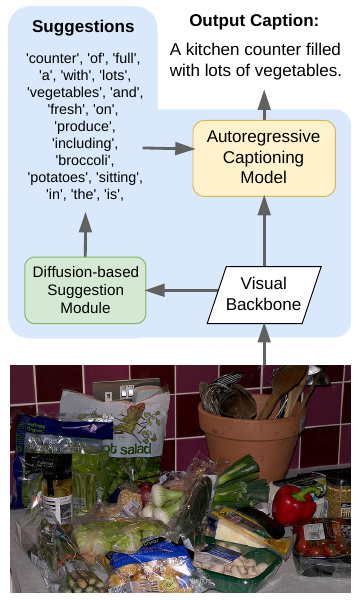}
  \caption{ \label{figure:first_page_pic} Our work proposes the adoption of diffusion models as an auxiliary tool that predicts suggestions for the autoregressive captioning model. The Figure reports a practical example of the idea.}
\end{figure}
work, we propose a new role for diffusion models, one that supports the autoregressive generation rather than replacing them. By doing so, we combine the best of the two worlds by leveraging the bidirectional processing capabilities of diffusion with the robust linguistic structure of the causal masking models to compensate for the loss of inter-token position dependency \cite{li2022multi, bao2021non}. 
To showcase the effectiveness of our solution, we present \textit{Show, Suggest and Tell (SST)}, a hybrid NAR/AR captioning model that achieves State-of-the-Art results on the COCO dataset, when trained on Cross-Entropy loss, outperforming most proposals in both diffusion and autoregressive modeling. 
Overall, the paper's contributions are the following: (i) We propose a new paradigm for diffusion models, with the adoption of diffusion models as Suggestion modules in the Image Captioning task. (ii) Based on the suggestion modules, we propose SST that achieves State-of-the-Art results on the MS-COCO 2014 test set, among similar models trained on Cross-Entropy loss. (iii) We perform extensive experiments to validate our proposal and showcase its generalizability across captioning models and diffusion strategies.

\section{Related Works}
\label{sec:related_works}

\subsection{Related Works in Standard Image Captioning}

\emph{Image Captioning With Auxiliary Modules.} Since the introduction of ``Show and Tell" \cite{tucker2012show}, one of the first Image Captioning Neural Networks, models have involved many intermediary mechanisms between the visual backbone and the decoding module. For instance, in ``Show, Attend, and Tell" \cite{xu2015show}, the attention mechanism guided the focus of each token toward a specific region of the convolution activation map. In ``Show, Observe and Tell" \cite{chen2018show}, in addition to the visual attention, they propose an additional Recurrent Neural Network that generates the image attributes such as adjectives and verbs.  In ``Show, Deconfound and Tell" \cite{Liu_2022_CVPR}, the author mitigates the presence of spurious correlations caused by visual and linguistic confounders in region-based image features with the help of causal graphs. Finally, in ``Show, Recall and Tell" \cite{wang2020show}, the authors proposed a Text-Retrieval module to collect a set of pertinent words that acted as a semantic guide or word-replacement during the text generation. In general, the adoption of auxiliary models is not a new concept in Image Captioning: the work of LTG \cite{jiang2018learning} 
proposed a guiding network that encapsulated the global visual information in each recurrent step, RL-MCTS \cite{luo2025clip} adopted CLIP contrastive model to guide the policy gradient learning based on Monte Carlo tree search.  
Recently, in \cite{alqatf2023guided}, the authors leveraged topic embeddings to convey context information more effectively, allowing them to generate more diverse and specific descriptions. Our proposal differs from established approaches in the adoption of a Diffusion model to provide suggestions to support the caption generation. 
Compared to the work of \cite{chen2018show}, our suggestion model incorporates a diffusion module and is trained on all reference words, not just the attributes. Additionally, in contrast to the recall mechanism, while suggestions are expected to be helpful and coherent, our architecture design actively prevents the copy mechanism from the AR model. 

\emph{Diffusion in Image Captioning.} Denoising Diffusion Models \cite{sohl2015deep, ho2020denoising} in Image Captioning mainly find application in data augmentation \cite{cioni2023diffusion, xiao2023multimodal} thanks to their generative capabilities, and in latency reduction \cite{zhou2021semi, liu2023prefix, guo2020non, xu2022clip, xiao2024diffusion} motivated by the ability to predict all tokens simultaneously. Several works have targeted the description quality specifically, such as DDCap \cite{zhu2022exploring}, SCD-Net \cite{luo2023semantic}, Ca-Captioner \cite{yang2024captioner}, and DECap \cite{wang2024decap}. Compared to these works, our proposal focuses on supporting autoregressive models rather than replacing them.

\subsection{Scope of the Work}

Given the explosion in popularity of Large Language Models (LLMs), we emphasize that our work focuses on architecture principles; as a result, our comparisons and proposal concentrate on models of standard size (around 100K parameters), trained on medium-sized datasets (COCO). We acknowledge the existence of many captioning proposals that significantly outperform the models presented in our papers, such as those trained on reinforcement learning \cite{rennie2017self, nguyen2022grit}, massive training corpora \cite{wang2022git}, or based on Vision Language Models such as \cite{zhou2020unified}. However, a proper stand-alone reinforcement learning for our proposal does not exist yet, and we believe it is impractical to do research on architecture principles with large language models or massive datasets without serious environmental impacts. Nonetheless, our work lays a groundwork that can be extended to Multimodal LLMs and massive datasets in the future. 


\begin{figure*}[ht]
  \centering  \includegraphics[width=0.85\textwidth]{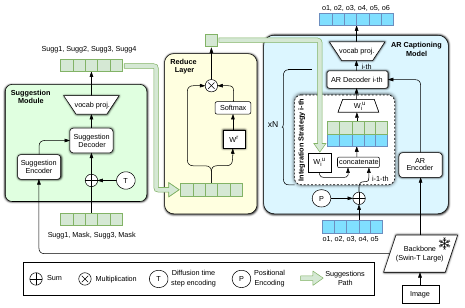}
  \caption{Show, Suggest, and Tell architecture. }
  \label{figure:architecture}
\end{figure*}

\section{Proposed Method}
\label{sec:method}

Our proposal mainly consists of two parts: a standard Autoregressive Neural Captioning architecture and a Discrete Diffusion Model, capable of Bidirectional Processing, that provides suggestion tokens to the former. Following, we provide the details for both components. 

\subsection{Suggestion Module}
\label{sec:sugg_module}

The $Suggestion \ Module$ consists of a Discrete Denoising Diffusion model, composed of a standard Transformer Encoder-Decoder architecture \cite{herdade2019image} on top of a pre-trained and frozen Visual Backbone, the Swin-Transformer Large \cite{liu2021swin}. As the architecture details are well known by the research community, we detail only the background and key aspects regarding the diffusion framework. 

Assuming $I$ is the input image, we can define $x_T(I)$ as a set of expected tokens to be found in a human description for the image $I$ (for the sake of visual clarity, we refer to $x_T(I)$ as $x_T$ in the following).  A Discrete Denoising Diffusion model (D3M) is characterized by a forward transition p($x_{t+1}|x_{t}$), which is a known probability function that applies noise to the input, generating multiple latent spaces of the target distribution, each one made of increasingly corrupted samples. For a sufficiently large selection of diffusion steps, the latent space corresponding to $x_0$ is known. Thus, if a Backward Transition function q($x_{t-1}|x_t$) was known, it could be used to generate samples from the target distribution. Since the mathematical definition of q($x_{t-1}|x_t$) is impractical, it is parametrized by a Neural Network $f_\theta(x_{t}, I)$ such that $q_{\theta}(x_{t-1}|x_{t}) = \mathcal{Q}(f_\theta(x_{t}, I))$ where $\mathcal{Q}$ denotes the overall denoising function. Among different possibilities for $\mathcal{Q}$, we adopt the reparametrized diffusion formulation of \cite{zheng2023reparameterized} and refer to their work for implementation details. Notably, our proposal and discussion are not bound to particular definitions of $\mathcal{Q}$ (See Section \ref{sec:diffusion_choice_ablation}). 

The proposed Suggestion Module covers the backward transition parametrization. We denote such function as $f^{sugg}_{\theta}(x_{t}, I)$ as an alias of the following operations:
\begin{equation}
\begin{aligned}
    L &= \text{Swin}\textnormal{-}\text{Transf}\textnormal{-}L(I) \\
    E &= \text{Enc}^{Sugg}(L) \\
    D &= \text{Dec}^{Sugg}(E, x_{t})
\end{aligned}
\label{eq:f_sugg}
\end{equation}
where $\text{Enc}^{Sugg}$ and $\text{Dec}^{Sugg}$ represent the stack of encoder and decoder layers respectively. Let $C(I)$ the set of human annotated caption for the image $I$, we define $x_0 = \{\{k_{j, i}\} \  \forall \ k_{j,i} \in c_i, \ \forall \ c_{i} \in C(I)\}$ as the set of unique tokens occurring in the set of human annotations. The Suggestion Module is trained to maximize the ELBO: 
\begin{equation}
\begin{aligned}
&\mathbb{E}_{p(x_1|x_0)}[log \ q_{\theta}(x_0|x_1)] - \\ \ 
\mathbb{E}_{p(x_t|x_0)}[&\mathcal{D}_{KL}(q(x_{t-1}|x_t, x_0) || q_{\theta}(x_{t-1}|x_t))]
\end{aligned}
\label{eq:f_sug_elbo}
\end{equation}
where $q_{\theta}(x_{t-1}|x_t) = \mathcal{Q}(f^{sugg}_{\theta}(x_{t}, I))$.

In addition to the denoising task, the Suggestion Module also includes a length decoder, which is responsible for predicting the number of suggested tokens. It presents the same architecture as the encoder, and it is placed on top of the visual features, and it is trained jointly with ELBO, on the following Length Loss (LL):
\begin{equation}
\begin{aligned}
    \text{Softmax}(W^{len} \text{AvgPool} ( \text{Encoder}^{len}(E_{N})))
\end{aligned}
\label{eq:f_sug_length}
\end{equation}
where $W^{len} \in \mathbb{R}^{H \times M}$, $M$ is the maximum amount of suggestions. During inference time, the length decoder is needed only once.

\subsection{Show Suggest and Tell (SST)}
\label{sec:show_suggest_tell}

The core of SST architecture consists of the adoption of the Suggestion Module and the integration strategy of the former predictions into the chosen $Autoregressive \ model$. In this work, we adopt the encoder-decoder refinement layers of ExpansionNet \cite{hu2023exploiting} as the neural captioner; notably, we report in Section \ref{sec:sugg_module_ablation} that our proposal is not limited to this choice.

Let $\text{Enc}\textnormal{-}\text{Layer}_{i}^{ExpNet}$ and $\text{Dec}\textnormal{-}\text{Layer}_{i}^{ExpNet}$ the $i-th$ encoder and decoder layer of ExpNet for $i = \{0, 1, ..., N\}$. The architecture and inner operations of the SST model are defined as follows:
\begin{equation}
\begin{aligned}
    E_0 &= \text{Swin}\textnormal{-}
    \text{Transf}\textnormal{-}L(I) \\
    D_0 &= Y_{l} \\
    \forall i &\in {1, 2, \ldots, N}\\
    &\begin{cases}
      E_i = \text{Enc}\textnormal{-}\text{Layer}_{i}^{ExpNet}(E_{i-1}) \\
       A_i = \text{Integration}\textnormal{-}\text{Module}(D_{i-1}, P) \\
      D_i = \text{Dec}\textnormal{-}\text{Layer}_{i}^{ExpNet}(E_N, A_{i-1}) \\
    \end{cases}  \\
    O_{l+1} &= \text{Softmax}(D_N) 
\end{aligned}
\label{eq:show_suggest_tell}    
\end{equation}
where $O_{l+1}$ is the vocabulary probability distribution for the prediction of the $l+1$ output token. Similarly,  $Y_{l}$ is the input sequence of the autoregressive generation at time step $l$. 
A core aspect of Equation \ref{eq:show_suggest_tell} is given by the Integration-Module, which is defined as:
\begin{equation}
\begin{aligned}
\text{Reduce} &= \text{Avg}(\text{Softmax}(PW^{r}) \odot P, axis=1) \\
A_i &= W_i^{u} \ [D_{i-1}, \text{Reduce}(P) ]_{axis=2}
\end{aligned}
\label{eq:att_reduce}
\end{equation}
where $P \in \mathbb{R}^{S \times H}$ are the embedding representations of the predictions generated by the Suggestion Module (see Section \ref{sec:sugg_module}), $[a,b]_{axis=2}$ denotes the concatenation operation over the hidden size, and $W_{i}^{r} \in \mathbb{R}^{H \times 1}, W_{i}^{u} \in \mathbb{R}^{2H \times H}$ are learned parameters. The $\odot$ symbolises a broadcasted product between the vector $PW_i^{r}$ and $P$. Overall, the $\text{Reduce} \ \text{Layer}$ generates a single feature vector consisting of a weighted average of all suggested tokens. Such a vector is then concatenated across all input sequences, denoted by $D_{i-1}$. In Equation \ref{eq:show_suggest_tell}, we omit for simplicity the basic components of diffusion and autoregressive models, such as time step encoding, positional encoding, embeddings, residual connections, and normalization and dropout layers. Such details will be found in the online repository. 

The autoregressive model is trained on the Standard Cross-Entropy: 
\begin{equation}
\begin{aligned}
 L_{XE}(\theta) = - \sum_{t}^{T} log( p_\theta(y^{*}_{t}|y^{*}_{1:t-1}, I))
\end{aligned}
\label{eq:xe_loss}
\end{equation}
where $p_\theta(y^{*}_{t}|y^{*}_{1:t-1}, I)$ the likelihood of $y^{*}_t$  given the previous words $y^{*}_{1:t-1}$.

The overall architecture, with each component depicted in Figure \ref{figure:architecture}. 

\begin{table*}[ht]
\centering
\caption{Comparison against State-of-the-Art architectures in Image Captioning on the MS-COCO 2014 test set. The highest score is highlighted in bold. }
\label{tab:sota_comparison}
\resizebox{0.85\textwidth}{!}{%
\begin{tabular}{llcccccccc}
\hline
Type & Model & Bleu1 & Bleu2 & Bleu3 & Bleu4 & Meteor & Rouge & Spice & Cider-D \\
\hline
{\multirow{12}{*}{Autoregressive}} & Show and Tell \cite{Vinyals_2015_CVPR} & - & - & - & 24.3 & 23.7 & - & - & 85.5 \\ 
{} & Show, Attend and Tell \cite{xu2015show} & 71.8 & 50.4 & 35.7 & 25.0 & 23.0 & - & - & - \\ 
{} & Show, Observe and Tell \cite{chen2018show} & 74.3 & 57.9 & 44.3 & 33.8 & - & 54.9 & - & 104.4  \\ 
{} & BUTD \cite{anderson2018bottom} &  77.2 & - & - & 36.2 & 27.0 & 56.4 & 20.3 & 113.5 \\ 
{} & ObjTrans \cite{herdade2019image} & 76.6 & - & - & 35.5 & 28.0 & 56.6 & 21.2 & 115.4 \\ 
{} & Transformer \cite{sharma2018conceptual}  & 76.1 & 59.9 & 45.2 & 34.0 & 27.6 & 56.2 & 21.0 & 113.3 \\ 
{} & Show, Recall and Tell \cite{wang2020show} & 77.1 & - & - & 36.6 & 28.0 & 56.9 & 21.3 & 116.9 \\ 
{} & CIIC$_{\mathcal{G}}$ \cite{Liu_2022_CVPR} & 77.5 & - & - & 37.3 & 28.5 & 57.4 & 21.5 & 119.0 \\ 
{} & AoA \cite{huang2019attention} & 78.7 & - & - & 38.1 & 28.5 & 58.2 & 21.7 & 122.7 \\ 
{} & X-Transformer \cite{pan2020x} & 77.3 & 61.5 & 47.8 & 37.0 & 28.7 & 57.5 & 21.8 & 120.0 \\ 
{} & PureT \cite{wang2022end} & 77.7 & 61.6 & 47.8 & 36.8 & 28.8 & 57.7 & 21.7 & 122.4 \\ 
{} & ExpansionNet v2 \cite{hu2023exploiting} & 77.1 & 61.2 & 47.6 & 36.6 & 29.1 & 57.8 & 22.7 & 123.6 \\ 
\hline
\multirow{6}{*}{Diffusion} & DDCap \cite{zhu2022exploring} & - & - & - & 34.7 & - & 58.0 & 21.5 & 116.7 \\ 
{} & BitDiffusion \cite{chen2022analog} & - & - & - & 34.7 & - & 58.0 & - & 115.0 \\
{} & SCD-Net \cite{luo2023semantic} & \textbf{79.0} & \textbf{63.4} & \textbf{49.1} & 37.3 & 28.1 & 58.0 & 21.6 & 118.0 \\
{} & DECap \cite{wang2024decap} & 78.5 & 62.2 & 47.4 & 35.3 & 29.0 & \textbf{58.4} & 22.7 & 121.2 \\
{} & Ca-Captioner \cite{yang2024captioner} & - & - & - & \textbf{38.3} & 28.8 & 57.7 & 21.8 & 122.0 \\
{} & EENAIC \cite{yu2023end} & 79.7 & - & - & 36.9 & 27.9 & 58.0 & - & 122.6 \\ 
\hline
\multirow{1}{*}{Mixed} & SST (Ours) & 78.3 & 62.6 & 48.8 & 37.6 & \textbf{29.2} & 58.3 & \textbf{22.8} & \textbf{125.1} \\ 
\hline
\end{tabular}
}
\end{table*}

\subsection{Design Motivation}
\label{sec:design_motivation}

\subsubsection{Suggestion Module} 
Our work focuses on showcasing the benefits of adopting diffusion models as suggestion modules rather than replacements of autoregressive counterparts. To this end, we kept the Suggestion Module architecture as simple as possible to showcase its effectiveness even in the most basic configuration. By doing so, we not only showcase the benefits of the proposal but also ensure that future research can easily extend our proposal with more sophisticated architectures.

\subsubsection{Show, Suggest and Tell} 
Since our proposal is not tied to a particular choice of an autoregressive network, the discussion reduces to the design of the integration module. We perform average-pooling of all suggested tokens into a single vector (Equation \ref{eq:att_reduce}). We argue that preventing the autoregressive model from accessing the complete suggestion tokens is preferred in the current state of research. Due to the limited capability of suggestion modules, if only a handful of correct but incomplete suggestions are provided, accessing all of them can act as noise or lead to an increase in exposure bias. Instead, by adopting a limited suggestion representation, we encourage the autoregressive model to adopt the suggestion as auxiliary information to guide the prediction, rather than a source that can be copied from. Our claim is supported by the experiments in Section \ref{sec:sugg_integration_strategy}.

\section{Experimental Results}
In the first Section \ref{sec:experimental_setup}, we detail the experimental details of the SST training, followed by a comparison against State-of-the-Art results in Section \ref{sec:comparison_sota}. In Sections \ref{sec:sugg_module}, \ref{sec:sugg_integration_strategy}, and \ref{sec:n_grams}, we report two ablation studies and discuss other suggestion strategies. Finally, in Section \ref{sec:qualitative_analysis} and \ref{sec:limits_suggestions}, we discuss the method's motivation and potential.

\subsection{Experimental Setup}
\label{sec:experimental_setup}

The autoregressive model of SST consists of three encoder-decoder layers and a hidden size of 512. The Suggestion Module is made of three layers, with a hidden size of 128. We train our proposal and ablation models on the popular MS-COCO dataset \cite{lin2014microsoft}. By adopting Karpathy's split, it results in 113,000 training image-description pairs, 5,000 for validation, and an equal amount for testing. 
The suggestion model is trained on a total of 565k human-annotated descriptions, five for each image. Each reference is pre-processed by lowering casing, removing punctuation, and filtering out words
that do not occur at least 5 times (vocabulary of size 10000). 

During the training, we adopt the learning rate and optimization configurations of \cite{hu2023exploiting} but train for 20 epochs instead. Additionally, to implement regularization, we provided suggestions to the captioning model only 50\% of the time (during inference, they are always offered). During inference time, autoregressive models perform beam search with a beam size of 3, whereas the Suggestion Modules perform a maximum of 20 diffusion steps. Captions are evaluated with BLEU \cite{papineni2002bleu}, ROUGE \cite{lin2004rouge}, CIDEr \cite{cioni2023diffusion}, METEOR \cite{denkowski2014meteor}, and SPICE \cite{anderson2016spice}, whereas suggestion proficiency is evaluated with standard Precision and Recall metrics. Precision is measured as the average on all images of $P=\#  \textnormal{correct suggestion} \ / \ \# \textnormal{suggestions} $. Similarly, Recall is computed as $R=\#\textnormal{correct suggestion} \ / \ |\mathcal{K}|$ where $|\mathcal{K}|$ is the number of unique tokens in the set of all human-annoted captions for each image.

\begin{table*}[ht]
\centering
\caption{Suggestion Module's Ablation Study on the MS-COCO 2014 validation sets.}
\label{tab:ablation_suggestion_F1}
\resizebox{0.9\textwidth}{!}{%
\begin{tabular}{clccccc}
\hline
Identifier &  Model Configuration & Training Configuration & Loss Function & Precision & Recall & F1 \\
\hline
(a) &  $\{H=512, L=3 \}$ & 1 Refs. per Image, All Words & ELBO + LL & 75.2\% & 27.2\% &  39.7\% \\
(b) &  $\{H=512, L=3 \}$ & 5 Refs. per Image, All Words & ELBO + LL & 58.7\% & 47.0\% & 51.7\% \\   
(c) & $\{H=128,L=3\}$ & 1 Refs per Image, All Words & ELBO + LL & 76.7\% & 26.9\% & 39.6\% \\ 
(d)\textsuperscript{*} & $\{H=128,L=3\}$ & 5 Refs per Image,  All Words & ELBO + LL & 62.0\% & 45.4\% & 51.8\% \\   
(e) & $\{H=128,L=3\}$ & 1 Refs per Image, Only Nouns \& Verbs & ELBO + LL & 72.4\% & 13.2\% &  22.2\% \\ 
(f) & $\{H=128,L=3\}$ & 5 Refs per Image, Only Nouns \& Verbs & ELBO + LL & 63.1\% & 22.0\% & 32.1\% \\     
(g) &  $\{H=128, L=3 \}$ & 1 Refs per Image, All Words & ELBO + LL + $A_{1H}$ & 77.9\% & 26.0\% & 38.7\% \\  
(h) & $\{H=128, L=3 \}$ & 5 Refs per Image, All Words & ELBO + LL + $A_{1H}$ & 61.0\% & 47.9\% & 53.2\% \\ 
\hline
\multicolumn{7}{l}{*: Adopted in Show, Suggest and Tell} \\
\end{tabular}
}
\end{table*}

\begin{table*}[ht]
\centering
\caption{Ablation Study of Show Suggest and Tell on the MS-COCO 2014 validation sets. Model identifiers refer to the naming of Table \ref{tab:ablation_suggestion_F1}.}
\resizebox{0.87\textwidth}{!}{%
\label{tab:ablation_suggestion_XE}
\begin{tabular}{lllccccccc}
\hline
Model & Suggestions & Bleu1 & Bleu2 & Bleu3 & Bleu4 & Meteor & Rouge & Spice & Cider-D \\
\hline
\multicolumn{10}{l}{Transformer with suggestions}  \\
\ \ \ \ Transformer Baseline & None & 75.3 & 59.1 & 45.4 & 34.7 & 28.6 & 56.6 & 21.9 & 117.0 \\        
\ \ \ \ \//w Suggestion Module (c) & $\checkmark$ & 76.5 & 60.6 & 46.8 & 35.7 & 28.9 & 57.4 & 22.2 & 119.2 \\        
\ \ \ \ \//w Suggestion Module (d) & $\checkmark$ & 77.7 & 61.9 & 48.0 & 36.8 & 28.8 & 57.7 & 22.3 & 120.6 \\
\ \ \ \ \//w Suggestion Module (g) & $\checkmark$ & 76.7 & 60.7 & 46.8 & 35.9 & 28.9 & 57.5 & 22.2 & 119.7 \\
\ \ \ \ \//w Suggestion Module (h) & $\checkmark$ & 77.0 & 61.0 & 47.2 & 36.5 &  29.2 & 57.9 & 22.3 & 120.9 \\

\multicolumn{10}{l}{}  \\
\multicolumn{10}{l}{Show, Suggest \& Tell Ablation}  \\
\ \ \ \ ExpansionNet v2 Baseline  & None & 77.5  & 62.0 & 48.4 & 37.5 & 29.3 & 58.1 & 22.6 & 123.3 \\
\ \ \ \ \//w Suggestion Module (c) & $\checkmark$ & 77.9 & 62.2 & 48.3 & 37.1 & 29.3 & 57.9 & 22.8 & 123.7 \\
\ \ \ \ \//w Suggestion Module (d)\textsuperscript{*} & $\checkmark$ & \textbf{78.8} & \textbf{63.5} & \textbf{49.5} & \textbf{38.1} & \textbf{29.4} & \textbf{58.6} & \textbf{22.7} & \textbf{124.9} \\ 
\ \ \ \ \//w Suggestion Module (g) & $\checkmark$ & 77.6 & 62.1 & 48.2 & 37.1 & 29.3 & 57.9 & 22.6 & 123.8 \\
\ \ \ \ \//w Suggestion Module (h) & $\checkmark$ & 78.0 & 62.5 & 48.8 & 37.7 & 29.4 & 58.2 & 22.6 & 124.5  \\
\hline
\multicolumn{10}{l}{*: Adopted in Show, Suggest and Tell} \\
\end{tabular}
}
\end{table*}

\subsection{Comparison Against Other Works}
\label{sec:comparison_sota}


We compare SST against other State-of-the-Art Image Captioning models trained on Cross-Entropy loss. For completeness, instances are selected from both Autoregressive and Diffusion-based models, as our proposal can be seen as a mixture of both strategies. Among all metrics, we focus on the CIDEr-D as it is the most comprehensive one. In Table \ref{sec:comparison_sota}, our proposal achieves 125.1 CIDEr-D, outperforms all the reported proposals in both autoregressive and diffusion cases by at least 1.5 and 2.5 points, respectively. However, we want to point out that our work likely does not hold the State-of-the-Art across all experimental setups, since we do not leverage reinforcement learning, a common practice in the field of Image Captioning. Due to the architectural focus of the work, such a comparison would not contribute to the scope and discussion of the paper. Instead, we highlight that, despite its simplicity, our proposal can outperform significantly both approaches in similar setups, not only redefining the role of diffusion models but also pointing toward a promising research direction.

\subsection{Suggestion Module Ablation Study}
\label{sec:sugg_module_ablation}

We compared the precision, recall, and F1 score for several configurations of Suggestion Modules. Several observations can be drawn from Table \ref{tab:ablation_suggestion_F1}. (i) Recalling that each image, in the case of COCO, is provided with five human-annotated captions, we note that grouping unique tokens from all reference captions is consistently more effective than training on one single caption per image. By observing each of the following pairs (a-b), (c-d), (e-f), and (g-h), the latter case exhibits higher precision but a significantly lower recall. While the higher precision can be attributed to lower prediction attempts, the lower recall showcases that the neural network suffers from forgetting. Such a result highlights the advantages of our proposed mixed approach, as the suggestion model can be aware of all references simultaneously, in contrast to the AR captioning model, where multiple references can conflict with each other. (ii) By comparing (a,b) and (c,d), we observe that a small hidden size of $H=128$ achieves the same performance as the popular selection of $H=512$. For this reason, we kept $H=128$ for all the following configurations. (iii) In the instances (d, e), we adjusted the suggestion problem by focusing only on nouns and verbs\footnote{To achieve that, we adopted Stanford's POS Tagger \cite{toutanova2003feature}
and we kept only the following tags ``NN", ``NNS", ``VBP", ``VBZ", ``CD", and ``VBG". }. In this case, we observed a significant drop in the recall score, suggesting that the most semantically meaningful words are also the most difficult to predict. This also suggests the infancy state of our proposed research direction, as the standard training frameworks and architecture may not be optimal for the suggestion problem. 
(iv) Finally, in (g, h), we extend the Set Average Cross Entropy ($A_{1H}$) \cite{asai2018set, dubuisson1994modified}, to sequences of sets, and integrate them in the diffusion loss. We define the $A_{1H}$ as follows: 
\begin{equation}
\begin{aligned}
    A_{1H}(X, Y) &= \frac{1}{|X|} \sum_{x \in X} min_{y \in Y} \ H(x,y) \\
\end{aligned}
\label{eq:a1h_for_sequences}
\end{equation}
where, $X \in \mathbb{R}^{T \times |V|}$, where $V$ is the vocabulary set and $T$ the sequence length.    $H(x,y)$ is the Cross-Entropy loss between the prediction distribution $x \in \mathbb{R}^{|V|}$ and target one-hot distribution $y = \mathbf{1_y} \in \mathbb{R}^{|V|}$. 

\begin{table*}[ht]
\centering
\caption{Suggestion Integration Comparison on the MS-COCO 2014 validation set.}
\label{tab:integration_strategy_ablation}
\begin{tabular}{clcccccccc}
\hline
Case & Integration & Bleu1 & Bleu2 & Bleu3 & Bleu4 & Meteor & Rouge & Spice & Cider-D \\
\hline
Baseline \cite{hu2023exploiting} & None & 77.5  & 62.0 & 48.4 & 37.5 & 29.3 & 58.1 & 22.6 & 123.3 \\
A & Cross-Attention over suggestions, in each decoder layer & 77.4 & 61.7 & 47.8 & 37.0 & 29.3 & 58.1 & 22.4 & 122.2 \\
B & Cross-Attention over suggestion, in each encoder layer & 77.5 & 61.9 & 48.2 & 37.3 & 29.3 & 58.3 & 22.4 & 122.8 \\
C & Reduce + Concatenation once, feed into encoder layers & 77.3 & 61.8 & 48.0 & 36.8 & \textbf{29.5} & 58.1 & \textbf{22.8} & 122.9 \\

D\textsuperscript{*} & Reduce + Concatenation once, feed into decoder layers & \textbf{78.6} & \textbf{63.5} & \textbf{49.5} & \textbf{38.1} & 29.4 & \textbf{58.6} & 22.7 & \textbf{124.9} \\
\hline
\multicolumn{10}{l}{*: Adopted in Show, Suggest and Tell.} \\
\end{tabular}
\end{table*}

Unfortunately, we observed only a slight improvement over instances (c, d). 
Since we did not observe significant differences in score with the hidden size selection and the integration of the adapted A\textsubscript{1H} loss, in our final SST model, we adopted the training configuration (d). Such a decision is further supported by the results in Table \ref{tab:ablation_suggestion_XE}, where we tested the impact of the most interesting configurations (c, d, g, h) on the caption generation. In this case, the following observations can be made. (i) We note that the methodology works on both the selected refining network ExpNet and the standard Transformer. This suggests the generalizability of our proposal. (ii) We observed a positive correlation between the F1 score and the increase in description quality (which can mostly be summarized by the CIDEr-D score). On this matter, since the case of A\textsubscript{1H} loss (h) did not yield a significant increase compared to the case (d), we did not include it in the final SST configuration.
Overall, our proposal introduces a noticeable increase in the caption quality, suggesting its effectiveness. 


\subsection{Suggestion Integration Strategy Ablation}
\label{sec:sugg_integration_strategy}

As discussed in Section \ref{sec:show_suggest_tell}, we adopted average-pooling of all suggested tokens into a single vector because we argue that, due to the limitations of the Suggestion Module, letting the AR captioning model have access to all tokens increases the exposure bias, as the network can be tempted in learning a copy-paste mechanism. Such a case is studied in Table \ref{tab:integration_strategy_ablation}, where the ``Reduce Layer" and integration modules are replaced by a Cross-Attention connected to all suggestions (all the remaining components and connections are preserved as in Figure \ref{figure:architecture}). In cases (A) and (B), the Cross-Attention connects the suggestions and the decoder and encoder, respectively, and it can be seen that both cases yield a decrease in the CIDEr-D score. Additionally, in the (C) case, we adopted the average pooling and concatenation approach, but fed the suggestions to the encoder layers instead. Such a setup proved to be ineffective, likely because suggestions act as a disturbance on the visual features. Overall, the results above confirmed our hypothesis and motivated our final design choices for the suggestion integration strategy (reported in the case (D)).

\begin{table*}[ht]
\centering
\caption{Diffusion Framework Comparison on the MS-COCO 2014 validation sets. All architectures follow a three-layer structure with a hidden dimension of 128.}
\label{tab:ablation_sugg_techniques}
\resizebox{0.9\textwidth}{!}{%
\begin{tabular}{lccccc}
\hline
Framework &  Training Configuration & Diffusion Loss Function & Precision & Recall & F1 \\
\hline

Direct Prediction (No Diffusion) & 5 Refs. per Image, All Words & Standard XE & 83.0\% & 26.3\% & 39.2\% \\
Absorbing Diffusion \cite{austin2021structured} & 5 Refs. per Image, All Words & ELBO  + LL & 62.7\% & 44.9\% &  51.6\% \\
Reparametrized Diffusion\textsuperscript{*} \cite{zheng2023reparameterized} & 5 Refs. per Image, All Words & ELBO + LL & 62.0\% & 45.4\% & 51.8\% \\
Analog-Bit Diffusion \cite{chen2022analog} & 5 Refs. per Image, All Words & ELBO + LL &  23.0\% & 82.6\% & 34.3\% \\
Analog-Bit Diffusion \cite{chen2022analog} & 5 Refs. per Image, All Words & ELBO + LL + LP & 47.6\% & 53.7\% &  49.9\% \\
\hline
\multicolumn{6}{l}{XE=Cross Entropy, LL=Length Loss, LP=Length Penalty.} \\
\multicolumn{6}{l}{*: Adopted in Show, Suggest and Tell.} \\
\end{tabular}
}
\end{table*}

\begin{table*}[ht]
\centering
\caption{Comparison between single words and n-grams on the suggestion  MS-COCO 2014 test sets.}
\label{tab:n_grams}
\begin{tabular}{lccccccccc}
\hline
Model & Suggestion Configuration & Bleu1 & Bleu2 & Bleu3 & Bleu4 & Meteor & Rouge & Spice & Cider-D \\
\hline
Baseline \cite{hu2023exploiting} & None & 77.5  & 62.0 & 48.4 & 37.5 & 29.3 & 58.1 & 22.6 & 123.3 \\ 
\multicolumn{10}{l}{\ \ \ \ \//w Suggestions}  \\
\ \ \ \ \//w Suggestions & N-gram=1, 1 Refs. per Image, All Words &  77.9 & 62.2 & 48.3 & 37.1 & 29.3 & 57.9 & 22.8 & 123.7 \\ 
\ \ \ \ \//w Suggestions$\textsuperscript{*}$ & N-gram=1, 5 Refs. per Image, All Words  &  78.3 & 62.6 & 48.8 & 37.6 & 29.2 & 58.3 & 22.8 & 125.1 \\ 
\ \ \ \ \//w Suggestions  & N-gram=2, 1 Refs. per Image, All Words   &  77.4 & 61.9 & 48.1 & 37.0 & 29.5 & 58.1 & 22.8 & 123.8 \\ 
\ \ \ \ "  & N-gram=2, 5 Refs. per Image, All Words  &  77.7 & 62.8 & 48.5 & 37.5 & 29.2 & 58.0 & 22.7 & 123.8 \\ 
\ \ \ \ "  & N-gram=3, 1 Refs. per Image, All Words  &  78.0 & 62.5 & 48.7 & 37.5 & 29.3 & 58.2  & 22.7 & 123.9 \\ 
\ \ \ \ " & N-gram=3, 5 Refs. per Image, All Words  &  78.3 & 62.9 & 49.1 & 37.8 & 29.3 & 58.3 & 22.8 & 124.2 \\ 
\hline
\multicolumn{10}{l}{*: Adopted in Show, Suggest and Tell.} \\
\end{tabular}
\end{table*}

\subsection{Impact of Diffusion Strategy on the Suggestion Module}
\label{sec:diffusion_choice_ablation}

To better motivate the adoption of a diffusion network, we showcase the impact of different prediction frameworks on the suggestion module. In particular, we focus on comparing three cases: the diffusion-less approach (Direct Prediction), the Analog-Bit Diffusion introduced by \cite{chen2022analog}, the Absorbing Diffusion \cite{austin2021structured}, and the one adopted by our proposal, the Reparametrized D3M proposed by \cite{zheng2023reparameterized}. 

First, it is worth noting that at least two approaches exist for the Direct Prediction approach. One consists of predicting all the tokens from the visual features, and the other one corresponds to the absorbing diffusion formulation with maximum diffusion steps of 1. To focus on generalizable principles, we only consider the latter case, as the former poses architectural constraints on the visual backbone and implies an intrinsic upper bound on the amount of suggested tokens. Table \ref{tab:ablation_sugg_techniques} confirms the superiority, hence the need for diffusion frameworks in contrast to a ``direct prediction" approach. In particular, the latter appears to yield only very safe predictions (e.g., mostly function words, resulting in high precision) and is incapable of generating complex suggestions, likely due to a lack of discriminative features in the input. This is solved by diffusion models thanks to the incremental token predictions. 

The absorbing \cite{austin2021structured} and our parametrized diffusion strategy \cite{zheng2023reparameterized} achieve the highest and most equilibrated results overall, motivating our final selection in SST. The case of vanilla Analog-Bit fundamentally differs from the other D3M in the lack of a length prediction. Instead, they are trained to denoise large vocabulary binary vectors whose value denotes the presence of a particular token in the final suggestion (in practice, we adopt Binary Cross-Entropy over each token in the vocabulary). As a result of noisy and long binary vectors, they tend to suggest many tokens (high recall), but many of them are incorrect (low precision). The issue was fixed with the addition of a length penalty loss defined by a Mean Squared Loss between the predictions accumulated across the vocabulary and the number of unique tokens. Thanks to this practice, the Analog-Bit diffusion achieved comparable scores to the other two. Overall, among diffusion approaches, we observe no significant differences between the three solutions. This indicates that our proposal is not tied to a particular diffusion and represents a generalizable principle in Neural Network designs.


\subsection{Single words versus N-grams}
\label{sec:n_grams}

Recalling that the Suggestion Module has access to all reference captions simultaneously, we examined whether considering only the set of unique tokens could limit its supporting capabilities. In particular, by training the auxiliary module to predict n-grams instead of tokens, we conducted a study about whether the captioning model could benefit from preserving the sentence's structural information in the suggestions. Unfortunately, as shown in Table \ref{tab:n_grams}, we observed that it was not the case, as the captioner performed similarly across all instances of n-gram size (for instance, N-gram=3 and N-gram=1 yield very similar results). We argue that while suggestions made of contiguous chunks of tokens can be effective, the increased complexity lowers the precision and recall to the point that it outweighs the benefits. 

\begin{figure}[ht]
  \centering
\includegraphics[width=0.48\textwidth]{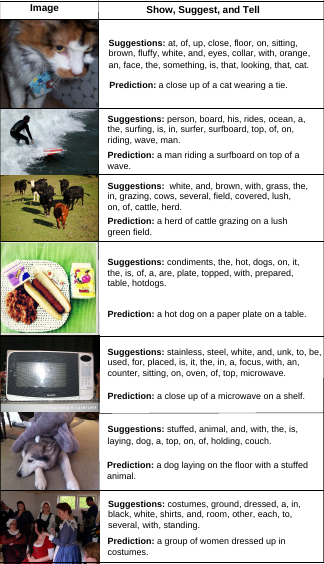}
  \caption{ \label{figure:example_suggestion} Suggestion examples of Show, Suggest, and Tell. Images are sampled from the validation set.}
\end{figure}


\subsection{Why It Works}
\label{sec:qualitative_analysis}

In Figure \ref{figure:example_suggestion}, some suggestions reported by SST are reported, and some observations can be drawn. It can be seen that most of the suggestions are correct, and even in cases where they are not included in the final prediction, it is often coherent with respect to the image. For instance, in the first image (from the top),  adjectives such as ``fluffy" and ``collar" 
do not appear in the final prediction, as traditional autoregressive models tend to be safer in the choice of words and ultimately less descriptive due to limited training samples. However, such adjectives can be freely predicted in the suggestion module. Ultimately, the Suggestion Module in SST can enforce a better understanding of the scenario by integrating an intermediary level of descriptions and enforcing additional text information, otherwise lost in standard autoregressive captioning\footnote{Since diffusion models access all tokens during the training, the proposal's effectiveness can be partially traced to the bidirectional induction principle \cite{hu2024bidirectional}.
}.

\subsection{Exploring The Potential of Suggestions}
\label{sec:limits_suggestions}

In this Section, we further highlight the importance of our work and empirically demonstrate why it is worth further investigation in future research. We adopt the same architecture of the SST module, but replace the suggestion module with an \textit{Oracle} that outputs portions of the ground truth during both the training and testing stages. Let $\mathcal{K}$ be the set of unique tokens. The output of Oracle's suggestion module is defined by randomly sampling $|K|$ times with probability $\rho \in [0, 1]$ from $K$, where $\rho$ is the preservation probability. For instance, for $\rho=1$, Oracle's suggestions correspond to $\mathcal{K}$.

In Table \ref{tab:limits_sugg}, we report the impact of different configurations for $\rho$ on the captioning model. Ideally, it represents how they can benefit from suggestions with an increasingly better suggestion module. As expected, there is a positive correlation between $\rho$ and the description quality. Most notably, in the case of BLEU4 and CIDEr-D, with a maximum of 40.1 and 132.6, respectively, at $\rho=1$. This result further confirms that the integration strategy discussed in Section \ref{sec:design_motivation} is indeed effective. For $\rho=0.25$, we observe that the Oracle outperforms our proposal, even though our suggestion module obtained an F1 score of 51.8\%. While such a discrepancy may sound odd at first, it actually matches the F1 score of SST when only nouns and verbs are considered in the computation, around 22\%. The discrepancy between Oracle $\rho=0.25$ and SST in performance can be explained by the former's capability to occasionally provide the right tokens in cases where our suggestion module struggles due to epistemic uncertainty. 
By this logic, results suggest that if the Suggestion Module can achieve a higher F1 score on nouns and verbs than 50\%, the captioning model can benefit enormously from it. This highlights the importance of our contribution, as our work taps into a fairly unexplored but promising research direction. 

\begin{table}[ht]
\centering
\caption{Show, Suggest, and Tell performance on MSCOCO 2014 test set with different selections of suggestion providers.}
\label{tab:limits_sugg}
\begin{tabular}{lcccccc}
\hline
Suggestions Provider & B1 & B4 & M & R & S & C \\
\hline
  Ours & 78.4 & 37.4 & 29.2 & 58.1 & 22.7 & 125.0 \\
  Oracle $\rho=0.25$  & 77.3 & 38.0 & 29.8 & 59.0 & 23.5 & 126.6 \\
  Oracle $\rho=0.50$ & 75.5 & 37.8 & 30.2 & 59.9 & 24.1 & 128.8 \\
  Oracle $\rho=0.75$ & 75.9 & 38.4 & 30.6 & 61.0 & 24.7 & 131.5 \\
  Oracle $\rho = 1.0$ & 77.6 & 40.1 & 31.0 & 61.4 & 25.3 & 132.6 \\
\hline
\end{tabular}
\end{table}

We remark that the results in Table \ref{tab:limits_sugg} are obtained using the simple integration strategy proposed in Section \ref{sec:method}, which was designed to prevent copy mechanisms. We believe that, in future works, if the suggestion module achieves better F1 scores, we can allow the autoregressive model to copy directly from suggestions, achieving even higher scores.

\section{Future Works}

Inspired by the suggestions reported in Figure \ref{figure:example_suggestion}, we observe that diffusion models can not only be effective as auxiliary tools, but also enable a new level of transparency. If the suggestion module is performing well enough, it can be used to monitor epistemic uncertainty, which is particularly useful in real-world applications. We believe this is a topic worth exploring in the future.
On top of that, in sections \ref{sec:n_grams} and  \ref{sec:limits_suggestions}, we argued that a stronger suggestion module can bring significant benefits to the caption's quality. This is further highlighted in Figure \ref{figure:example_suggestion}. While most suggestions are accurate and coherent, two limitations can be observed: (i) It appears that some images receive more suggestions than others, suggesting that the suggestion module, similar to any Neural Network, might suffer from an unbalanced dataset or linguistic biases in descriptions. (ii) In some cases, even the most apparently obvious suggestions are missing, such as ``women" in the last image's example (from the top). This latter issue is likely caused by the infancy stage of the research topic, since the discrete diffusion framework was originally designed for sequences, which is \textit{not} the case of Equation.  Finally, our work focuses on a fundamental level by proposing a new architecture principle. Thus, by incorporating larger datasets, in future works, it can be extended to Large Language Models (LLMs), which are predominantly made of autoregressive models. Our work provides a solution to the lack of presence of diffusion models in the LLM scene. 

\section{Conclusion}

In this work, we propose the adoption of diffusion models as auxiliary models to help standard captioning, rather than challenging the status quo. In particular, we presented the ``Show, Suggest and Tell (SST)" architecture, which is equipped with a Suggestion Module that learn words it expects to find in the final prediction and feeds them to the autoregressive model. By doing so, the model can explicitly predict additional attributes and objects to enforce the understanding of the scenery, even when the suggested tokens are not included in the final description. Results on the COCO dataset showcase that our joint proposal outperforms established works of both autoregressive and diffusion models. Extensive experiments are reported to validate the proposed principle and showcase its significance. Finally, the simplicity of our method, combined with the observed positive outcomes, points to a fairly untapped potential and a promising direction for future research, in terms of both captioning improvements and explainable AI.

\section*{Acknowledgment}

This work has received funding from the European Union’s Horizon 2020 programme dAIEDGE (G.A. No 101120726). This work has also been supported by the CINECA Italian consortium, project ISCRA-ShareGPT.

\bibliographystyle{abbrv}
\bibliography{IEEE-conference-template-062824/my_references}

\end{document}